\documentclass[a4paper, 10pt, conference]{ieeeconf}        % Use this line for a4
\usepackage{FG2021}

\FGfinalcopy % *** Uncomment this line for the final submission

\IEEEoverridecommandlockouts                               % This command is only
\usepackage{graphicx} % for pdf, bitmapped graphics files
\usepackage{epsfig} % for postscript graphics files
\usepackage{mathptmx} % assumes new font selection scheme installed
\usepackage{times} % assumes new font selection scheme installed
\usepackage{amsmath} % assumes amsmath package installed
\usepackage{amssymb}  % assumes amsmath package installed
\usepackage{booktabs} % To thicken table lines
\usepackage{multirow} % join rows in tables

\usepackage{url}

\def\FGPaperID{****} % *** Enter the FG2021 Paper ID here

\title{\LARGE \bf
Action Transformer: A Self-Attention Model for Short-Time Pose-Based Human Action Recognition
}

\author{\parbox{16cm}{\centering
    {\large Vittorio Mazzia$^{1,2,3}$, Simone Angarano$^{1,2}$, Francesco Salvetti$^{1,2,3}$, Federico Angelini$^4$ and Marcello Chiaberge$^{1,2}$}\\
    {\normalsize
    $^1$ Department of Electronics and Telecommunications, Politecnico di Torino, 10124 Turin, Italy\\
     $^2$ PIC4SeR, Politecnico di Torino Interdepartmental Centre for Service Robotics, Turin, Italy\\
     $^3$ SmartData@PoliTo, Big Data and Data Science Laboratory, Turin, Italy\\
     $^4$ School of Computing, Newcastle University, 1 Science Square, Newcastle, NE45TG, UK}}
    \thanks{This work has been developed with the contribution of the Politecnico di Torino Interdepartmental Centre for Service Robotics PIC4SeR (https://pic4ser.polito.it) and SmartData@Polito (https://smartdata.polito.it).}% <-this % stops a space
}

\begin{document}

\ifFGfinal
\thispagestyle{empty}
\pagestyle{empty}
\else
\author{Anonymous FG2021 submission\\ Paper ID \FGPaperID \\}
\pagestyle{plain}
\fi
\maketitle

%%%%%%%%%%%%%%%%%%%%%%%%%%%%%%%%%%%%%%%%%%%%%%%%%%%%%%%%%%%%%%%%%%%%%%%%%%%%%%%%
\begin{abstract}
Deep neural networks based purely on attention have been successful across several domains, relying on minimal architectural priors from the designer. In Human Action Recognition (HAR), attention mechanisms have been primarily adopted on top of standard convolutional or recurrent layers, improving the overall generalization capability. In this work, we introduce Action Transformer (AcT), a simple, fully self-attentional architecture that consistently outperforms more elaborated networks that mix convolutional, recurrent and attentive layers. In order to limit computational and energy requests, building on previous human action recognition research, the proposed approach exploits 2D pose representations over small temporal windows, providing a low latency solution for accurate and effective real-time performance. Moreover, we open-source MPOSE2021, a new large-scale dataset, as an attempt to build a formal training and evaluation benchmark for real-time, short-time HAR. The proposed methodology was extensively tested on MPOSE2021 and compared to several state-of-the-art architectures, proving the effectiveness of the AcT model and laying the foundations for future work on HAR. 
\end{abstract}

%%%%%%%%%%%%%%%%%%%%%%%%%%%%%%%%%%%%%%%%%%%%%%%%%%%%%%%%%%%%%%%%%%%%%%%%%%%%%%%%
\section{INTRODUCTION} % simo
\label{intro}

\begin{figure}[t]
    \includegraphics[width=\columnwidth]{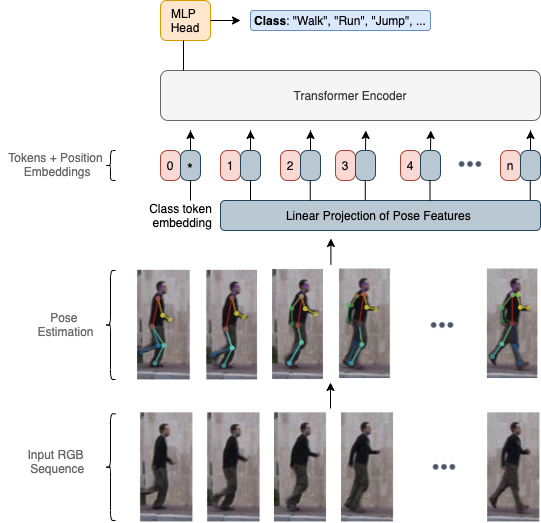}
    \caption{Overview of the Action Transformer architecture. Pose estimations are linearly projected to the dimension of the model, and together with the class token, they form the input tokens of the transformer encoder. As for Vision Transformer models \cite{dosovitskiy2020image, touvron2020training,pmlr-v139-d-ascoli21a}, a learnable positional embedding is added to each input token. Then, only the output class token is passed through a multi-layer perceptron head to obtain the final class prediction.}
    \label{fig:act_arch}
\end{figure}

Human Action Recognition (HAR) is a problem in computer vision and pattern recognition that aims to detect and classify human actions. The ability to recognize people inside a scene and predict their behavior is fundamental for several applications, such as robotics \cite{rodriguez2020shedding}, surveillance and security \cite{Vallathan2021, xuanming2021}, autonomous vehicles \cite{martin2019drive, ben2021driving}, and automatic video captioning \cite{zhou2018end, tu2021enhancing, wan2022revisiting}. Most previous works dealing with HAR adopted datasets characterized by samples with a long temporal duration \cite{HUANG2019165, 7940083, luvizon2020multi}. Thus, HAR has been mainly treated as a post-processing operation, classifying complex and long-lasting human actions by exploiting past and future information. Conversely, in this work, we focus on short-time HAR, which aims at continuously classifying actions within short past time steps (up to a second). This approach is fundamental to target real-time applications: in robotics, for example, HAR problem should be solved promptly to react to sudden behavioral changes, only relying on near past information.

In this paper, we propose a new model for HAR called the Action Transformer (AcT), schematized in Fig.\ref{fig:act_arch}, inspired by the simple and prior-free architecture of the Vision Transformer \cite{dosovitskiy2020image}. The Transformer architecture \cite{vaswani2017attention} has been one of the most important deep learning advances of the last years in natural language processing (NLP). Moreover, multi-head self-attention has proven to be effective for a wide range of tasks besides NLP, e.g. image classification \cite{dosovitskiy2020image,touvron2020training,pmlr-v139-d-ascoli21a}, image super-resolution \cite{Yang_2020_CVPR, zhu2020learning}, and speech recognition \cite{8462506}. Furthermore, optimized versions of the Transformer have been developed for real-time and embedded applications \cite{berg2021keyword}, proving that this architecture is also suitable for Edge AI purposes.
Recently, many models for Human Action Recognition have proposed the integration of attention mechanisms with convolutional and recurrent blocks to improve the accuracy of models. However, solutions that rely exclusively on self-attention blocks have not been investigated for this task yet.

With AcT, we apply a pure Transformer encoder derived architecture to action recognition obtaining an accurate and low-latency model for real-time applications. We study the model at different scales to investigate the impact of the number of parameters and attention heads. We use the new single-person, short-time action recognition dataset MPOSE2021 as a benchmark and exploit 2D human pose representations provided by two existing detectors: OpenPose \cite{cao2019openpose} and PoseNet \cite{papandreou2018personlab}. Moreover, we compare AcT with other state-of-the-art baselines to highlight the advantages of the proposed approach. To highlight the effectiveness of self-attention, we conduct a model introspection providing visual insights of the results and study how a reduction of input temporal sequence length affects accuracy. We also conduct extensive experimentation on model latency on low-power devices to verify the suitability of AcT for real-time applications.

The main contributions of this work can be summarized as follows:
\begin{itemize}
    \item We study the application of the Transformer encoder to 2D pose-based HAR and propose the novel AcT model, proving that fully self-attentional architectures can outperform existing convolutional and recurrent models for pose-based HAR.
    \item We introduce MPOSE2021, a dataset for real-time short-time HAR, suitable for both pose-based and RGB-based methodologies. It includes 15429 sequences from 100 actors and different scenarios, with limited frames per scene (between 20 and 30). In contrast to other publicly available datasets, the peculiarity of having a constrained number of time steps stimulates the development of real-time methodologies that perform HAR with low latency and high throughput.
\end{itemize}

The rest of the paper is organized as follows. Section \ref{relatedworks} briefly reports the relevant research on HAR and self-attention deep learning methodologies. Section \ref{mpose2021} presents the MPOSE2021 dataset, highlighting features and elements of novelty with respect to other existing datasets. Then, Section \ref{actiontransformer} describes the architecture of the proposed AcT model, shortly recalling the main features of the Transformer encoder. Section \ref{experiments} summarizes the experimentation conducted to verify the effectiveness of the proposed approach, gives a visual insight into the model functioning, studies its behavior under temporal information reduction, and measures its latency for real-time applications. Section \ref{conclusion} draws some conclusions for this work and indicates future research directions.

%%%%%%%%%%%%%%%%%%%%%%%%%%%%%%%%%%%%%%%%%%%%%%%%%%%%%%%%%%%%%%%%%%%%%%%%%%%%%%%%
\section{RELATED WORKS}
\label{relatedworks}
\begin{figure}[b]
    \centering
    \includegraphics[width=\linewidth]{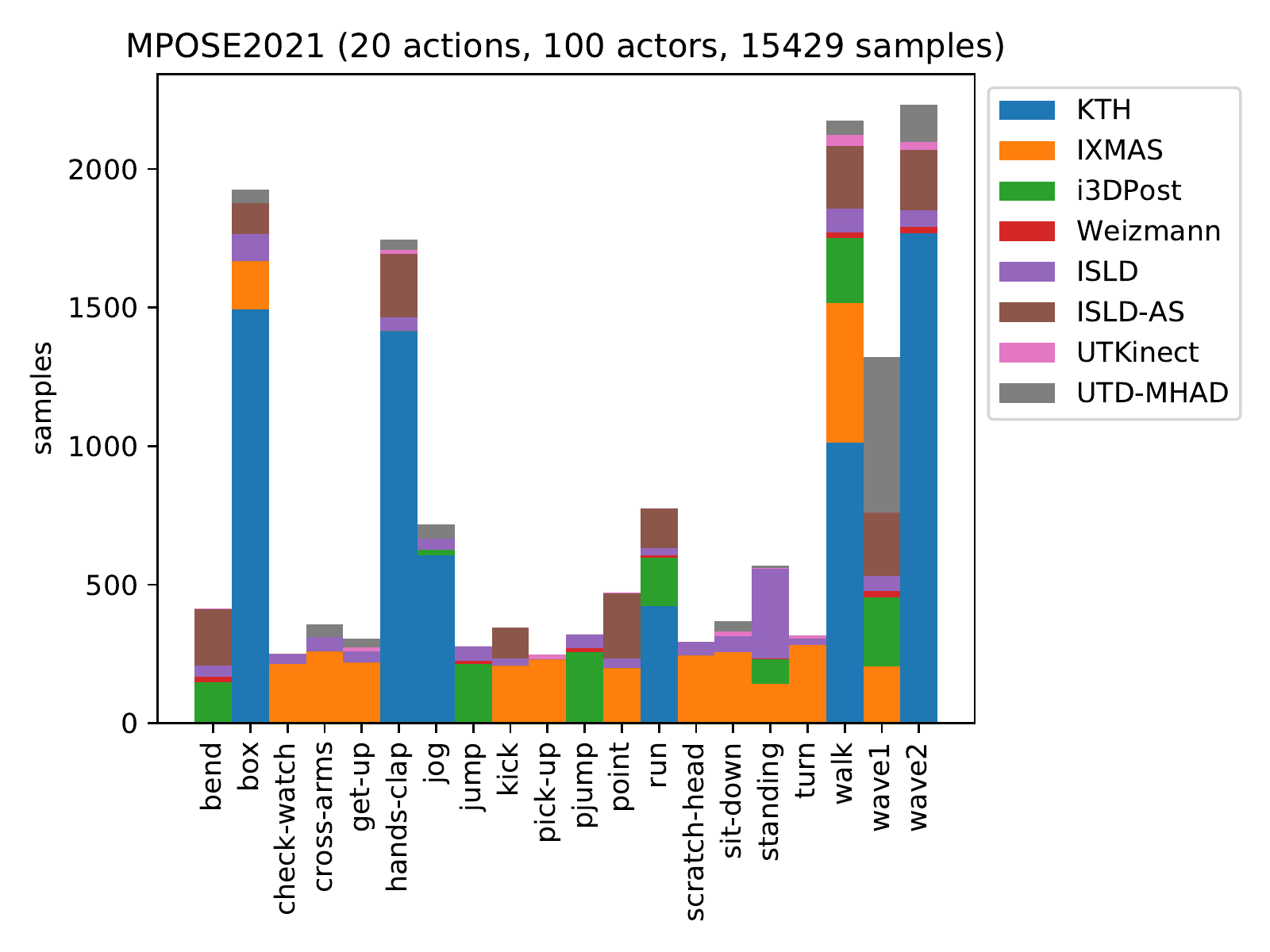}
    \caption{The number of samples of MPOSE2021 divided by action. The colors show the distribution of precursor datasets among classes, highlighting the unbalanced nature of the data. The final dataset contains 15429 samples where each sample represents one of 100 actors performing one of 20 actions.}
    \label{fig:mpose2021-summary}
\end{figure}
Human Action Recognition algorithms aim at detecting and classifying human behaviors based on a different source of information. Works in this field can be mainly subdivided into two broad categories: video-based and depth-based methodologies\cite{song2021human}. In both, the input consists of a sequence of points representing body joints, extracted from the corresponding RGB frame\cite{LI2020107037}, in the case of video-based methods, or a point cloud, in the case of depth-based ones. In the latter, body joints are provided as 3D coordinates (skeletal data), such as those captured by Kinect sensors \cite{shahroudy2016ntu,liu2019ntu}, that can be possibly projected onto the 2D space \cite{kay2017kinetics,yan2018spatial}. On the opposite, in this work, we focus on a video-based analysis, where body joints are already provided as 2D coordinates (pose data) by pose detection algorithms such as OpenPose \cite{cao2019openpose} and PoseNet \cite{papandreou2018personlab}. This characteristic makes 2D HAR methodologies applicable in a great range of applications using a simple RGB camera. Conversely, skeletal data require particular sensors to be acquired, such as Kinect or other stereo-cameras. That raises substantial limitations, such as availability, cost, limited working range (up to 5-6 meters in the case of Kinect \cite{langmann2012depth}), and performance degradation in outdoor environments. 

Angelini et al.\ first collected the MPOSE dataset for video-based short-time HAR \cite{angelini2018actionxpose}, obtaining 2D poses by processing several popular HAR video datasets with OpenPose. Moreover, the authors proposed ActionXPose, a methodology that increases model robustness to occlusions. The sequences were classified using MLSTM-FCN \cite{karim2018multi}, which exploits a combination of 1D convolutions, LSTM \cite{hochreiter1997long}, and Squeeze-and-excitation attention \cite{hu2018squeeze}. The same authors successively applied their approach to anomaly detection \cite{angelini2019privacy,angelini2019joint} and expanded MPOSE with the novel ISLD and ISLD-Additional-Sequences datasets. Conversely, Yan et al. \cite{yan2018spatial} first applied OpenPose to extract 2D poses from the Kinetics-400 RGB dataset \cite{kay2017kinetics} and used graph convolutions to capture spatial and temporal information. 
Similarly, Shi et al. \cite{shi2019skeleton,shi2019two} applied graph convolutions to pose information using two streams that extract information from both joints and bones to represent the skeleton. Hao et al.\cite{HAO201913} introduced knowledge distillation and dense-connectivity to explore the spatiotemporal interactions between appearance and motion streams along different hierarchies. At the same time, Si et al. \cite{SI2020107511} put together a hierarchical spatial reasoning network and temporal stack learning network to extract discriminative spatial and temporal features.
On the other hand, Liu et al. \cite{liu2020disentangling} proposed an ensemble of two independent bone-based and joint-based models using a unified spatial-temporal graph convolutional operator. Conversely, Cho et al. \cite{cho2020self} first applied self-attention \cite{vaswani2017attention} to the skeletal-based HAR problem. More recently, Plizzari et al. \cite{plizzari2021skeleton}, inspired by Bello et al. \cite{bello2019attention}, employed self-attention to overcome the locality of the convolutions, again adopting a two-stream ensemble method, where self-attention is applied on both temporal and spatial information.

Unlike previous methodologies, this paper presents an architecture for HAR entirely based on the Transformer encoder, without any convolutional or recurrent layer. Moreover, we focus on verifying the suitability of our method for low-latency and real-time applications. For this reason, we introduce a new 2D pose-based dataset specifically designed for short-time HAR. Traditional datasets used by previous works, such as 3D NTU RGB+D \cite{shahroudy2016ntu,liu2019ntu}, or Kinetics-Skeleton \cite{kay2017kinetics,yan2018spatial}, include long temporal sequences which must be entirely scanned to make the correct classification. In contrast, the proposed MPOSE2021 dataset includes samples with a temporal duration of at most 30 frames. That makes it a new and more suitable benchmark to test the short-time and low-latency performance of HAR models.

%%%%%%%%%%%%%%%%%%%%%%%%%%%%%%%%%%%%%%%%%%%%%%%%%%%%%%%%%%%%%%%%%%%%%%%%%%%%%%%%
\section{THE MPOSE2021 DATASET} % federico
\label{mpose2021}
\begin{figure}[t]
    \centering
    \includegraphics[width=\linewidth]{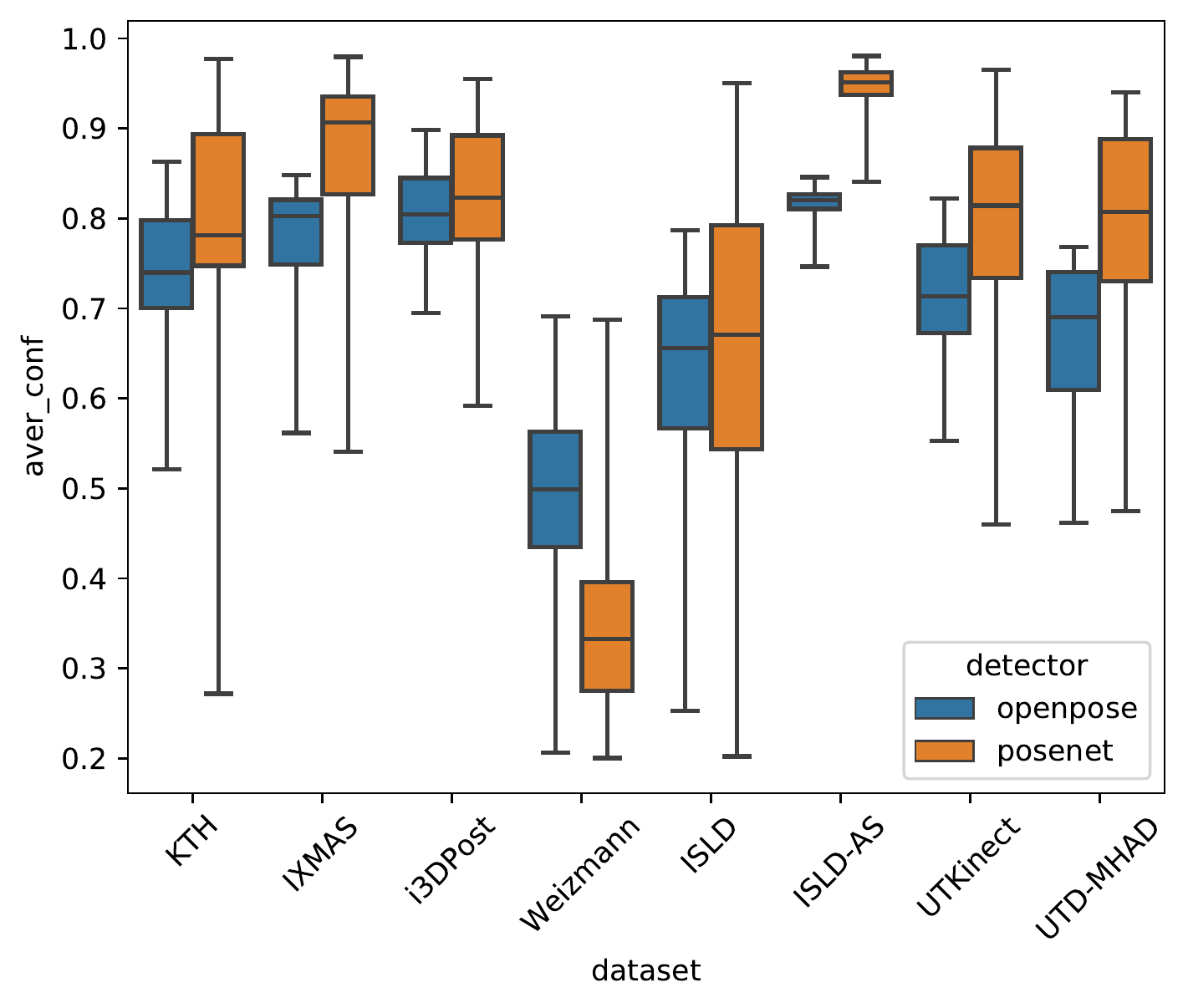}
    \caption{Comparison between OpenPose and PoseNet average confidence (aver\_conf) in the different MPOSE2021 sub-datasets. The detectors achieve different average confidence based on the considered precursor dataset.}
    \label{fig:mpose2021-aver-conf}
\end{figure}
In this section, MPOSE2021 is presented as an RGB-based dataset designed for short-time, pose-based HAR. As in \cite{angelini2018actionxpose, angelini2019privacy, angelini2019joint}, video data have been previously collected from popular HAR datasets (precursors), i.e. Weizmann \cite{gorelick2007actions}, i3DPost \cite{gkalelis2009i3dpost}, IXMAS \cite{weinland2006free}, KTH \cite{schuldt2004recognizing}, UTKinetic-Action3D (RGB only) \cite{xia2012view}, UTD-MHAD (RGB only) \cite{chen2015utd}, ISLD, and ISLD-Additional-Sequences \cite{angelini2018actionxpose}. 

\begin{figure}[t]
    \centering
    \includegraphics[width=\linewidth]{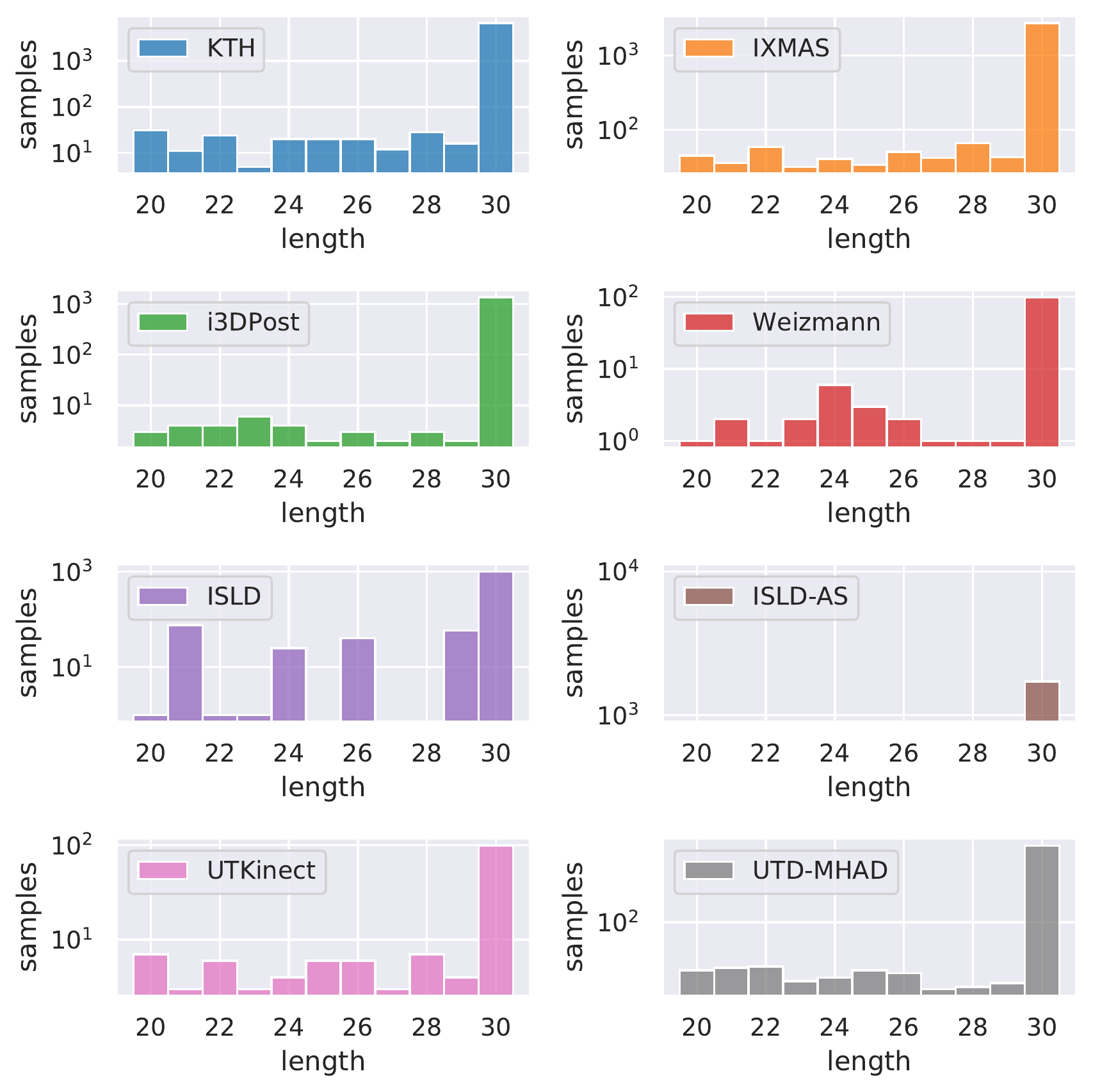}
    \caption{MPOSE2021 sequence length distribution for the different sub-datasets. It is possible to notice how most of them present a mode of 30 frames per scene.}
    \label{fig:mpose2021-frame-num}
\end{figure}

Due to the heterogeneity of actions across different datasets, labels are remapped to a list of 20 common classes. Actions that cannot be remapped accordingly are discarded. Therefore, precursor videos are divided into non-overlapping samples (clips) of 30 frames each whenever possible and retaining tail samples with more than 20 frames. A visual representation of the number of frames per sample for each sub-dataset of MPOSE2021 is shown in Fig. \ref{fig:mpose2021-frame-num}. The peculiarity of a reduced number of time steps contrasts with other publicly available datasets and stimulates the development of methodologies that require low latency to perform a prediction. That would largely benefit many real-world applications requiring real-time perception of the actions performed by humans nearby.

Subsequently, clips not containing a single action are discarded. Moreover, ambiguous clips are relabelled whenever possible or discarded otherwise. This process leads to 15429 samples, where each sample represents a single actor performing a single action. The total number of distinct actors in MPOSE2021 is 100, and the number of samples for each action is reported in Fig.\ref{fig:mpose2021-summary}, which also shows the distribution of precursor datasets.

OpenPose \cite{cao2019openpose} and PoseNet \cite{papandreou2018personlab} are used to extract landmarks from MPOSE2021 samples. The average confidence is computed for each sample as the mean across landmarks and frames. It turns out that the two detectors achieve different average confidences based on the considered precursor dataset. The box plot of Fig.\ref{fig:mpose2021-aver-conf} describes the comparison statistics.

Due to the significant sample heterogeneity and the high number of actors, three different training/testing splits are defined for MPOSE2021, i.e. Split1, Split2, and Split3, by randomly selecting 21 actors for testing and using the rest of them for training. This division makes the proposed dataset a challenging benchmark to effectively assess and compare the accuracy and robustness of different methodologies. Moreover, the suggested evaluation procedure requires testing a target model on each split using ten validation folds and averaging the obtained results. That makes it possible to produce statistics and reduces the possibility of overfitting the split testing set with an accurate choice of hyperparameters.

With MPOSE2021, we aim at providing an end-to-end and easy-to-use benchmark to robustly compare state-of-the-art methodologies for the short-time human action recognition task. We thus release a code repository\footnote{https://github.com/PIC4SeRCentre/MPOSE2021} to access the different levels of the dataset (video, RGB frames, 2D poses). Moreover, we open source a practical Python package to access, visualize and preprocess the poses with standard functions. The Python package can be easily installed with the command \hspace{0.02cm} \texttt{pip install mpose}.

%%%%%%%%%%%%%%%%%%%%%%%%%%%%%%%%%%%%%%%%%%%%%%%%%%%%%%%%%%%%%%%%%%%%%%%%%%%%%%%%
\section{ACTION TRANSFORMER} % vitto
\label{actiontransformer}
In this section, we describe the architecture of the AcT network (Fig.\ref{fig:act_arch}) briefly recalling some preliminary concepts associated with the Transformer model\cite{vaswani2017attention}.

\subsection{AcT Architecture}
A video input sequence with $T$ frames of dimension $H \times W$ and $C$ channels $\textbf{X}_{rgb} \in \mathbb{R}^{T \times H \times W \times C}$ is pre-processed by a multi-person 2D pose estimation network
\begin{equation}
    \textbf{X}_{2Dpose} = F_{2Dpose}(\textbf{X}_{rgb})
\end{equation}
that extracts 2D poses of dimension $N \times T \times P$, where $N$ is the number of human subjects present in the frame and $P$ is the number of keypoints predicted by the network. 
The Transformer architecture receives as input a 1D sequence of token embeddings, so each of the $N$ sequences of pose matrices $\textbf{\textit{X}}_{2Dpose} \in  \mathbb{R}^{T \times P}$ is separately processed by the AcT network. Nevertheless, at inference time, all detected poses in the video frame can be batch-processed by the AcT model, simultaneously producing a prediction for all $N$ subjects.
\begin{figure}[t]
    \includegraphics[width=\linewidth]{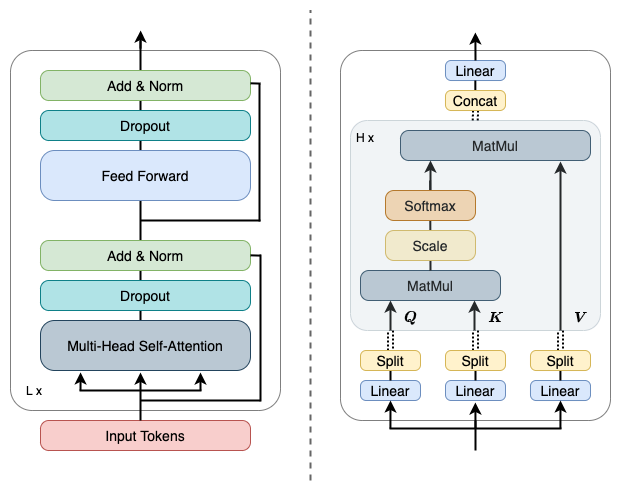}
    \caption{Transformer encoder layer architecture (left) and schematic overview of a multi-head self-attention block (right). Input tokens go through L encoder layers and H self-attention heads.}
    \label{fig:act_inside}
\end{figure}
Firstly, the $T$ poses are mapped to a higher dimension $D_{model}$ using a linear projection map $\textbf{\textit{W}}^{l_{0}} \in \mathbb{R}^{P \times D_{model}}$. As in BERT \cite{devlin2018bert} and Vision Transformers \cite{dosovitskiy2020image, touvron2020training, berg2021keyword, caron2021emerging}, a trainable vector of dimension $D_{model}$ is added to the input $T$ sequence. This class token [CLS] forces the self-attention to aggregate information into a compact high-dimensional representation that separates the different action classes. Moreover, positional information is provided to the sequence with a learnable positional embedding matrix $\textbf{\textit{X}}_{pos} \in \mathbb{R}^{(T+1)\times D_{model}}$ added to all tokens.

The linearly projected tokens and [CLS] are fed to a standard Transformer encoder $F_{Enc}$ of L layers with a post-norm layer normalization \cite{vaswani2017attention, chen2018best}, obtaining

\begin{equation}
    \textbf{\textit{X}}^{L} = F_{Enc}(\textbf{\textit{X}}^{l_{0}}) = F_{Enc}([\textbf{\textit{x}}_{cls}^{l_0};\textbf{\textit{X}}_{2Dpose}] + \textbf{\textit{X}}_{pos})
\end{equation}
where $\textbf{\textit{X}}^{L}\in \mathbb{R}^{(T+1) \times D_{model}}$ is the overall representation produced by the Transformer encoder at its last layer. Finally, only the [CLS] token $\textbf{\textit{x}}_{cls}$ is fed into a linear classification head $\text{MLP}_{head}$ that performs the final class prediction

\begin{equation}
    \hat{\textbf{\textit{z}}} = \text{MLP}_{head}(\textbf{\textit{x}}_{cls}^{L})
\end{equation}

where $\hat{\textbf{\textit{z}}}$ is the output logit vector of the model. At training time, the supervision signal comes only from the [CLS] token, while all remaining $T$ tokens are the only input of the model. It is important to notice how the nature of the network makes it possible to accept a reduced number of frames as input even if trained with a fixed $T$. That gives an additional degree of freedom at inference time, making AcT more adaptive than other existing models.

The resulting network is a lightweight solution capable of predicting actions for multiple people in a video stream with high accuracy. The advantage of building on 2D pose estimations enables effective real-time performance with low latency and energy consumption.

\subsection{Transformer Architecture}
The Transformer encoder \cite{vaswani2017attention} is made of L layers with alternating H multi-head self-attention and feed-forward blocks. Dropout \cite{srivastava2014dropout}, Layernorm \cite{ba2016layer}, and residual connections are applied after every block. The overall sequence of blocks of a Transformer encoder is summarized on the left of Fig.\ref{fig:act_inside}.

Each feed-forward block is a multi-layer perceptron with two layers and GeLu \cite{hendrycks2016gaussian} non-linearity. The first layer expands the dimension from $D_{model}$ to $D_{mlp}=4\cdot D_{model}$ and applies the activation function. On the other hand, the second layer reduces the dimension back from $D_{mlp}$ to $D_{model}$.  

Instead, the multi-head $\textbf{\textit{Q}}\textbf{\textit{K}}\textbf{\textit{V}}$ self-attention mechanism (MSA) is based on a trainable associative memory with key-value vector pairs.
For the l-th layer of the Transformer encoder and the h-th head, queries ($\textbf{\textit{Q}}$), keys ($\textbf{\textit{K}}$) and values ($\textbf{\textit{V}}$) are computed as $\textbf{\textit{Q}}=\textbf{\textit{X}}\textbf{\textit{W}}_Q$, $\textbf{\textit{K}}=\textbf{\textit{X}}\textbf{\textit{W}}_K$ and $\textbf{\textit{V}}=\textbf{\textit{X}}\textbf{\textit{W}}_V$ respectively, where $\textbf{\textit{W}}_Q$, $\textbf{\textit{W}}_K$ and $\textbf{\textit{W}}_V$ belong to $\mathbb{R}^{D_{model}\times D_{h}}$ , being $D_{h}$ the dimension of the attention head. So, for each self-attention head (SA) and element in the input sequence $\textbf{\textit{X}} \in \mathbb{R}^{T \times D_{model}}$ we compute a weighted sum over all values $\textbf{\textit{V}}$. The attention weights $A_{ij}$ are derived from the pairwise similarity between two elements of the sequence and their respective query $Q_i$ and key $K_j$ representations. Therefore, it is possible to compute the attention weights $\textbf{\textit{A}} \in \mathbb{R}^{T \times T}$ for the l-th layer as

\begin{table}[t]
\caption{Action Transformer parameters for the four version sizes. We fix $D_{model}/\text{H}=64$, linearly increasing H, $D_{mlp}$, and L in order to obtain different versions of the AcT network.}
\centering
\begin{tabular}{cccccc}
\toprule
\textbf{Model} & H & $D_{model}$ & $D_{mlp}$ & L & Parameters \\ \hline 
AcT-$\mu$         & 1 & 64       & 256    & 4 & 227k          \\ 
AcT-S         & 2 & 128      & 256    & 5 & 1,040k        \\ 
AcT-M         & 3 & 192      & 256    & 6 & 2,740k        \\ 
AcT-L         & 4 & 256      & 512    & 6 & 4,902k        \\ 
\bottomrule
\end{tabular}
\label{tab:settings}
\end{table}

\begin{equation}
    \textbf{\textit{A}}=\text{Softmax}\left(\frac{\textbf{\textit{Q}}\textbf{\textit{K}}^{T}}{\sqrt{D_{h}}}\right)
\end{equation}
and the final weighted sum SA as
\begin{equation}
    \text{SA}(\textbf{\textit{X}})=\textbf{\textit{A}}\textbf{\textit{V}}
\end{equation}
We perform this SA operation for all the H heads of the l-th layer. We concatenate the results and linearly project the output tensor to the original dimension model as
\begin{equation}
    \text{MSA}(\textbf{\textit{X}})=[SA_{1}(\textbf{\textit{X}});SA_{2}(\textbf{\textit{X}});...;SA_{\text{H}}(\textbf{\textit{X}})]\textbf{\textit{W}}_{MSA}
\end{equation}

where $\textbf{\textit{W}}_{MSA} \in \mathbb{R}^{\text{H}\cdot D_{h}\times D_{model}}$.  All operations are schematized on the right side of Fig. \ref{fig:act_inside}. The input tensor $\textbf{\textit{X}}$ is firstly projected to $T\times \text{H} \cdot D_{h}$, and then a SA operation is performed to all H splits of the resulting projected tensor. Finally, all head outputs are concatenated and linearly projected back to $D_{model}$. So, the attention mechanism takes place in the time-domain, allowing a global class embedding representation by correlating different time windows.

In order to reduce the number of hyperparameters and linearly scale the dimension of AcT, we fix $D_{model}/\text{H}=64$, varying H, $D_{mlp}$, and L to obtain different versions of the network. A simple grid search using train and validation sets is performed to determine lower and upper bounds for the four parameters. In particular, in Table \ref{tab:settings}, we summarize the four AcT versions with their respective number of parameters. The four models (micro, small, medium, and large) differ for their increasing number of heads and layers, substantially impacting the number of trainable parameters.

%%%%%%%%%%%%%%%%%%%%%%%%%%%%%%%%%%%%%%%%%%%%%%%%%%%%%%%%%%%%%%%%%%%%%%%%%%%%%%%%
\section{EXPERIMENTS}
\label{experiments}
This section describes the main experiments conducted to study the advantages of using a fully self-attentional model for 2D pose-based HAR. First, the four variants of AcT described in Section \ref{actiontransformer} are compared to existing state-of-the-art methodologies and baselines on MPOSE2021. Only for a specific comparison with ST-TR \cite{plizzari2021skeleton} and MS-G3D \cite{liu2020disentangling}, we use additional ensemble versions of our model named AcT-$\mu$ (x $n$), $n$ being the number of ensembled instances. Then, we further analyze the behavior of the network in order to get a visual insight of the attention mechanism and study the performance under a reduction of temporal information. Finally, we study model latency for all the designed architectures with two different CPU types, proving that AcT can easily be used for real-time applications.

\subsection{Experimental Settings}
In the following experiments, we employ both the OpenPose and PoseNet versions of the MPOSE2021 dataset. Either set of data has $T=30$ and $P=52$ or $68$ features, respectively. In particular, for OpenPose we follow the same preprocessing of \cite{angelini2018actionxpose} obtaining 13 keypoints with four parameters each (position $x,y$ and velocities $v_x,v_y$), while PoseNet samples contain 17 keypoints with the same information. The training set is composed of 9421 samples and 2867 for testing. The remaining 3141 instances are used as validation to find the most promising hyperparameters with a grid search analysis. All results, training, and testing code for the AcT model are open source and publicly available\footnote{https://github.com/PIC4SeRCentre/AcT}.

\begin{table}[t]
\scriptsize
\caption{Hyperparameters used in AcT experiments.}
\label{tab:hyperparams_act}
\centering
\begin{tabular}{ll}
\toprule
\multicolumn{2}{c}{\textbf{Training}} \\
\midrule
Training epochs & 350 \\
Batch size & 512 \\
Optimizer & AdamW \\
Warmup epochs & 40\% \\
Step epochs & 80\% \\
\bottomrule
\end{tabular}
\qquad
\begin{tabular}{ll}
\toprule
\multicolumn{2}{c}{\textbf{Regularization}} \\
\midrule
Weight decay & 1e-4 \\
Label smoothing & 0.1 \\
Dropout & 0.3 \\
Random flip & 50\% \\
Random noise $\sigma$ & 0.03 \\
\bottomrule
\end{tabular}
\end{table}

\begin{table*}[ht]
\centering
\caption{Benchmark of different models for short-time HAR on MPOSE2021 splits using OpenPose 2D skeletal representations.}
\begin{tabular}{@{}cccccccc@{}}
\toprule
\multicolumn{2}{c}{\textbf{MPOSE2021 Split}} & \multicolumn{2}{c}{\textbf{OpenPose 1}} & \multicolumn{2}{c}{\textbf{OpenPose 2}} & \multicolumn{2}{c}{\textbf{OpenPose 3}} \\ \midrule
\textbf{Model} & \textbf{Parameters} & \textbf{Accuracy {[}\%{]}} & \textbf{Balanced {[}\%{]}} & \textbf{Accuracy {[}\%{]}} & \textbf{Balanced {[}\%{]}} & \textbf{Accuracy {[}\%{]}} & \textbf{Balanced {[}\%{]}} \\ \midrule
\textbf{MLP} & 1,334k & 82.66 ± 0.33 & 74.56 ± 0.56 & 84.41 ± 0.60 & 74.58 ± 1.00 & 83.48 ± 0.58 & 76.60 ± 0.77 \\
\textbf{Conv1D} & 4,037k & 88.18 ± 0.64 & 81.97 ± 1.40 & 88.93 ± 0.43 & 80.49 ± 0.95 & 88.67 ± 0.38 & 83.93 ± 0.58 \\
\textbf{REMNet \cite{angarano2021robust}} & 4,211k & 89.18 ± 0.51 & 84.20 ± 0.84 & 88.77 ± 0.35 & 80.29 ± 0.88 & 89.80 ± 0.59 & 86.18 ± 0.40 \\ \midrule
\textbf{ActionXPose \cite{angelini2018actionxpose}} & 509k & 87.60 ± 0.98 & 82.13 ± 1.50 & 88.42 ± 0.70 & 81.28 ± 1.40 & 89.96 ± 1.00 & 86.65 ± 1.60 \\
\textbf{MLSTM-FCN \cite{karim2018multi}} & 368k & 88.62 ± 0.74 & 83.55 ± 0.88 & 90.19 ± 0.68 & 83.84 ± 1.20 & 89.80 ± 0.94 & 87.33 ± 0.67 \\ 
\textbf{T-TR \cite{plizzari2021skeleton}} & 3,036k & 87.72 ± 0.87 & 81.99 ± 1.64 & 88.14 ± 0.53  & 80.23 ± 1.19  & 88.69 ± 0.95 & 85.03 ± 1.60 \\
\textbf{MS-G3D (J) \cite{liu2020disentangling}} & 2,868k & 89.90 ± 0.50 & 85.29 ± 0.98 & 90.16 ± 0.64 & 83.08 ± 1.10 & 90.39 ± 0.44 & 87.48 ± 1.20 \\

\midrule
\textbf{AcT-$\mu$} & 227k & 90.86 ± 0.36 & 86.86 ± 0.50 & 91.00 ± 0.24 & 85.01 ± 0.51 & 89.98 ± 0.47 & 87.63 ± 0.54 \\
\textbf{AcT-S} & 1,040k & 91.21 ± 0.48 & 87.48 ± 0.76 & 91.23 ± 0.19 & 85.66 ± 0.58 & 90.90 ± 0.87 & 88.61 ± 0.73 \\
\textbf{AcT-M} & 2,740k & \textbf{91.38 ± 0.32} & \textbf{87.70 ± 0.47} & 91.08 ± 0.48 & 85.18 ± 0.80 & 91.01 ± 0.57 & 88.63 ± 0.51 \\
\textbf{AcT-L} & 4,902k & 91.11 ± 0.32 & 87.27 ± 0.46 & \textbf{91.46 ± 0.42} & \textbf{85.92 ± 0.63} & \textbf{91.05 ± 0.80} & \textbf{89.00 ± 0.74} \\
\midrule
\textbf{ST-TR \cite{plizzari2021skeleton}} & 6,072k	& 89.20 ± 0.71 & 83.95 ± 1.11 & 89.29 ± 0.81 & 81.53 ± 1.39 & 90.49 ± 0.53 & 87.06 ± 0.70\\
\textbf{MS-G3D (J+B) \cite{liu2020disentangling}} & 5,735k & 91.13 ± 0.33 & 87.25 ± 0.50 & 91.28 ± 0.29 & 85.10 ± 0.50 & 91.42 ± 0.54 & 89.66 ± 0.55 \\

\textbf{AcT-$\mu$ (x2)} & 454k & 91.76 ± 0.29 & 88.27 ± 0.37 & 91.34 ± 0.40 & 86.88 ± 0.48 & 91.70 ± 0.57 & 88.87 ± 0.37 \\
\textbf{AcT-$\mu$ (x5)} & 1,135k & 92.43 ± 0.24 & 89.33 ± 0.31 & 91.55 ± 0.37 & 87.80 ± 0.39 & 92.63 ± 0.55 & 89.77 ± 0.35 \\
\textbf{AcT-$\mu$ (x10)} & 2,271k & \textbf{92.54 ± 0.21} & \textbf{89.79 ± 0.34} & \textbf{92.03 ± 0.33} & \textbf{88.02 ± 0.31} & \textbf{93.10 ± 0.53} & \textbf{90.22 ± 0.31} \\
\bottomrule  
\end{tabular}
\label{tab:benchmark_openpose}
\end{table*}

\begin{figure*}[ht]
    \centering
    \includegraphics[width=\textwidth]{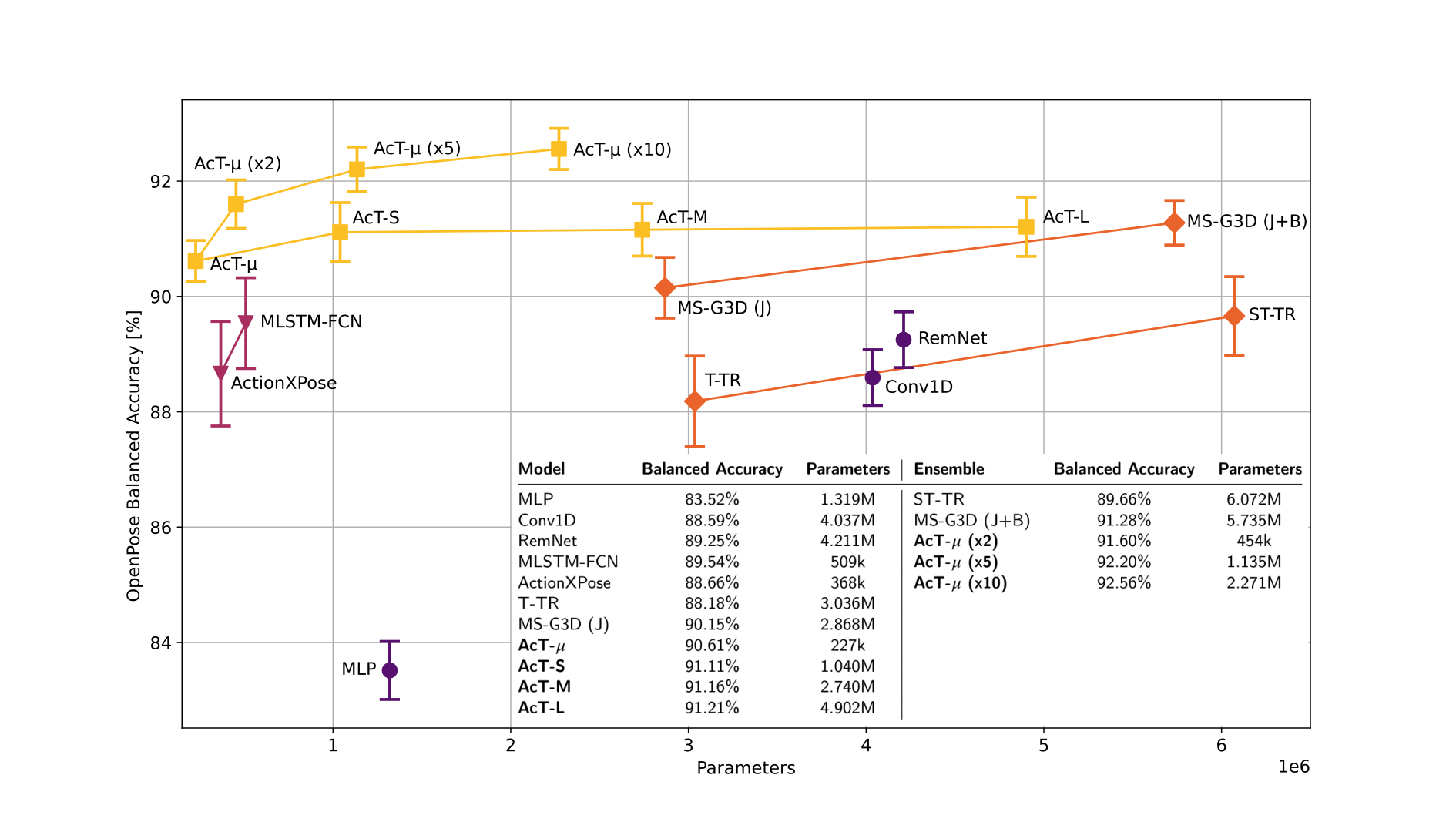}
    \caption{Visual representation of the benchmark of different models for short-time HAR on MPOSE2021 splits using OpenPose 2D skeletal representations. For brevity and clearness, the average balanced accuracy on the three splits of MPOSE2021 is reported. The lines connect models that use the same methodology.}
    \label{fig:mean_balanced}
\end{figure*}

In Table \ref{tab:settings} are summarized all the hyperparameters for the four versions of the AcT architecture and in Table \ref{tab:hyperparams_act} all settings related to the training procedure.
The AdamW optimization algorithm \cite{loshchilov2017decoupled} is employed for all training with the same scheduling proposed in \cite{vaswani2017attention}, but with a step drop of the learning rate $\lambda$ to 1e-4 at a fixed percentage (80\%) of the total number of epochs. We employ the TensorFlow 2\footnote{https://www.tensorflow.org} framework to train the proposed network on a PC with 32-GB RAM, an Intel i7-9700K CPU, and an Nvidia 2080 Super GP-GPU. Following the previously defined benchmark strategy, the total training procedure for the four versions takes approximately 32 hours over the three different splits. We exploit publicly available code for what concerns other state-of-the-art models and use the same hyperparameters and optimizer settings described by the authors in almost all the cases. The only exception is made for learning rate, epochs number, and batch size to adapt the methodologies to our dataset and obtain better learning curves. 

\begin{table*}[ht]
\centering
\caption{Benchmark of different models for short-time HAR on MPOSE2021 splits using PoseNet 2D skeletal representations.}
\begin{tabular}{@{}cccccccc@{}}
\toprule
\multicolumn{2}{c}{\textbf{MPOSE2021 Split}} & \multicolumn{2}{c}{\textbf{PoseNet 1}} & \multicolumn{2}{c}{\textbf{PoseNet 2}} & \multicolumn{2}{c}{\textbf{PoseNet 3}} \\ \midrule
\textbf{Model} & \textbf{Parameters} & \textbf{Accuracy {[}\%{]}} & \textbf{Balanced {[}\%{]}} & \textbf{Accuracy {[}\%{]}} & \textbf{Balanced {[}\%{]}} & \textbf{Accuracy {[}\%{]}} & \textbf{Balanced {[}\%{]}} \\ \midrule
\textbf{Conv1D} & 4,062k & 85.83 ± 0.71 & 79.96 ± 1.10 & 87.47 ± 0.35 & 78.51 ± 0.78 & 87.46 ± 0.67 & 81.31 ± 0.58 \\
\textbf{REMNet \cite{angarano2021robust}} & 4,269k & 84.75 ± 0.65 & 77.23 ± 0.94 & 86.17 ± 0.68 & 75.79 ± 1.30 & 86.31 ± 0.60 & 79.20 ± 0.79 \\ \midrule
\textbf{ActionXPose \cite{angelini2018actionxpose}} & 509k & 75.98 ± 0.72 & 64.47 ± 1.10 & 79.94 ± 1.10 & 67.05 ± 1.40 & 77.34 ± 1.40 & 66.86 ± 1.40\\
\textbf{MLSTM-FCN \cite{karim2018multi}} & 368k & 76.17 ± 0.84 & 64.75 ± 1.10 & 79.04 ± 0.72 & 65.62 ± 1.40 & 77.84 ± 1.30 & 67.05 ± 1.20 \\

\midrule
\textbf{AcT-$\mu$} & 228k & 86.66 ± 1.10 & 81.56 ± 1.60 & 87.21 ± 0.99 & 79.21 ± 1.60 & 87.75 ± 0.53 & 82.99 ± 0.87 \\
\textbf{AcT-S} & 1,042k & \textbf{87.63 ± 0.52} & \textbf{82.54 ± 0.87} & 88.48 ± 0.57 & 81.53 ± 0.68 & 88.49 ± 0.65 & 83.63 ± 0.99 \\
\textbf{AcT-M} & 2,743k & 87.23 ± 0.48 & 82.10 ± 0.66 & \textbf{88.50 ± 0.51} & \textbf{81.79 ± 0.44} & \textbf{88.70 ± 0.57} & \textbf{83.92 ± 0.96} \\
\bottomrule
\end{tabular}
\label{tab:benchmark_posenet}
\end{table*}

\subsection{Action Recognition on MPOSE2021}
We extensively experiment on MPOSE2021 considering some baselines, common HAR architectures, and our proposed AcT models. We report the mean and standard deviation of 10 models trained using different validation splits to obtain statistically relevant results. The validation splits are constant for all the models and correspond to 10\% of the train set, maintaining the same class distribution. The benchmark is executed for both OpenPose and PoseNet data and repeated for all the three train/test splits provided by MPOSE2021. The baselines chosen for the benchmark are a Multilayer Perceptron (MLP), a fully convolutional model (Conv1D), and REMNet, which is a more sophisticated convolutional network with attention and residual blocks proposed in \cite{angarano2021robust} for time series feature extraction. In particular, the MLP is designed as a stack of three fully connected (FC) layers with 512 neurons each, followed by a dropout layer and a final FC layer with as many output nodes as classes. Instead, the Conv1D model is built concatenating five 1D convolutional layers with 512 filters alternated with batch normalization stages and followed by a global average pooling operator, a dropout layer, and an FC output stage as in the MLP. Finally, the configuration used for REMNet consists of two Residual Reduction Modules (RRM) with a filter number of 512, followed by dropout and the same FC output layer as in the other baselines.

Regarding state-of-the-art comparisons, four popular models used for multivariate time series classification, and in particular HAR, are reproduced and tested. Among those, MLSTM-FCN \cite{karim2018multi} combines convolutions, spacial attention, and an LSTM block, and its improved version ActionXPose \cite{angelini2018actionxpose} uses additional preprocessing, leading the model to exploit more correlations in data and, hence, be more robust against noisy or missing pose detections. On the other hand, MS-G3D \cite{liu2020disentangling} uses spatial-temporal graph convolutions to make the model aware of spatial relations between skeleton keypoints, while ST-TR \cite{plizzari2021skeleton} joins graph convolutions with Transformer-based self-attention applied to both space and time. As the last two solutions also propose a model ensemble, these results are further compared to AcT ensembles made of 2, 5, and 10 single-shot models. We also report the achieved balanced accuracy for each model and use it as the primary evaluation metric to account for the uneven distribution of classes. 

The results of the experimentation for OpenPose are reported in Table \ref{tab:benchmark_openpose} and, in synthesis, in Fig.\ref{fig:mean_balanced}. The fully convolutional baseline strongly outperforms the MLP, while REMNet proves that introducing attention and residual blocks further increases accuracy. As regards MLSTM-FCN and ActionXPose, it is evident that explicitly modeling both spatial and temporal correlations, made possible by the two separate branches, slightly improves the understanding of actions with respect to models like REMNet. MS-G3D, in its joint-only (J) version, brings further accuracy improvement by exploiting graph convolutions and giving the network information on the spatial relationship between keypoints. On the other hand, ST-TR shows performance comparable to all other single-shot models, despite, as MS-G3D, taking advantage of graph information.

The proposed AcT model demonstrates the potential of pure Transformer-based architectures, as all four versions outperform other methodologies while also showing smaller standard deviations. Moreover, even the smallest AcT-$\mu$ (227k parameters) is able to extract general and robust features from temporal correlations in sequences. Increasing the number of parameters, a constant improvement in balanced accuracy can be observed for splits 3, while split 1 and 2 present oscillations. The difference between splits reflects how much information is conveyed by training sets and how much the model can learn from that. So, it is evident that AcT scales best on split three because it presents complex correlations that a bigger model learns more easily. On the contrary, it seems AcT-$\mu$ is able to extract almost all the information relevant for generalization from split 2, as the accuracy only slightly increases going towards more complex models.

\begin{figure*}[t]
    \centering
    \includegraphics[width=\textwidth]{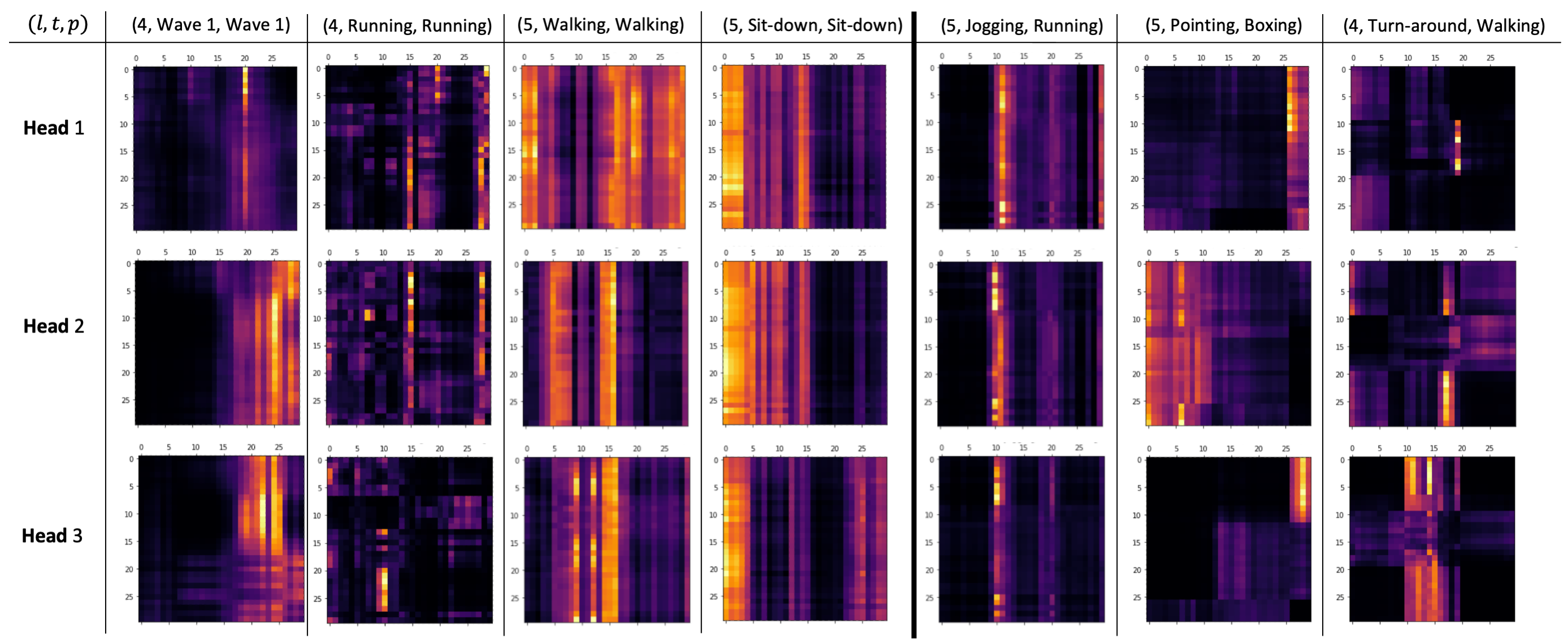}
    \caption{Self-attention weights $\textbf{\textit{A}}$ of MPOSE2021 test samples. $(l,t,p)$ represents the AcT-M l-th layer, the true label and the prediction respectively. The three rightmost columns show three attention maps of a failed prediction and the other columns are from correct classifications. It is clear from all examples how the model focuses on certain particular frames of the series in order to extract a global representation of the scene.}
    \label{fig:attention_vis}
\end{figure*}

As explained in Section \ref{relatedworks}, ST-TR exploits an ensemble of two networks modeling spatial and temporal sequence correlations, respectively. Moreover, MS-G3D leverages further information such as skeleton graph connections and the position of bones in one of the ensembled networks. Ensembles are very effective in reducing model variance and enhancing performance by exploiting independent representations learned by each network, so it is unfair to compare them with single architectures. For this reason, we create three ensemble versions of AcT-$\mu$ to have an even confront, with 2, 5, and 10 instances, respectively. To compute ensemble predictions, we average the output logits of the network instances before passing through a softmax function. The results reported at the bottom of Table \ref{tab:benchmark_openpose} show that AcT-$\mu$ (x2) outperforms MS-G3D (J+B) in all the benchmarks except for balanced accuracy in split 3, despite having less than one-tenth of its parameters. Finally, the ensembles AcT-$\mu$ (x5) and AcT-$\mu$ (x10), made of 5 and 10 instances respectively, achieve even higher accuracy on all the splits with only around 1 to 2 million parameters. That proves how the balancing effect of the ensemble enhances model predictions even without feeding the network additional information.

As PoseNet data is mainly dedicated to real-time and Edge AI applications, only the models designed for this purpose have been considered in the benchmark, excluding MS-G3D, ST-TR, and AcT-L. In general, the results give similar insights. The tested models are the same as the previous case, with all the necessary modifications given by the different input formats. In the MLP case, however, performance seriously degrades as networks strongly tend to overfit input data after a small number of epochs, so the results are not included in Table \ref{tab:benchmark_posenet}. That is caused by the fact that PoseNet is a lighter methodology developed for Edge AI, and hence noisy and even missing keypoint detections are more frequent. That results in less informative data and emphasizes the difference between sequences belonging to different sub-datasets, confusing the model and inducing it to learn very specific but unusable features. The MLP is too simple and particularly prone to this kind of problem. Naturally, all the models are affected by the same problem, and the balanced accuracy on PoseNet is generally lower. The same considerations made for OpenPose apply in this case, where AcT outperforms all the other architectures and demonstrates its ability to give an accurate and robust representation of temporal correlations. Also, it is interesting to notice that Conv1D performs better than REMNet, proving to be less prone to overfitting, and that standard deviations are more significant than in the OpenPose case.

\subsection{Model Introspection} % vitto 
In order to have an insight into the frames of the sequence the AcT model attends to, we extract the self-attention weights at different stages of the network. In Fig.\ref{fig:attention_vis}, MPOSE2021 test samples are propagated through the AcT-M model, and attention weights $\textbf{\textit{A}}$ of the three distinct heads are presented. It can be seen that the model pays attention to specific frames of the sequence when a specific gesture defines the action. On the other hand, attention is much more spread across the different frames for more distributed actions such as walking and running. Moreover, it is clear how the three heads mostly focus on diverse frames of the sequence. Finally, the rightmost columns show three attention maps of failed predictions. In these last cases, attention weights are less coherent, and the model cannot extract a valid global representation of the scene.

\begin{figure}[t]
    \centering
    \includegraphics[width=\columnwidth]{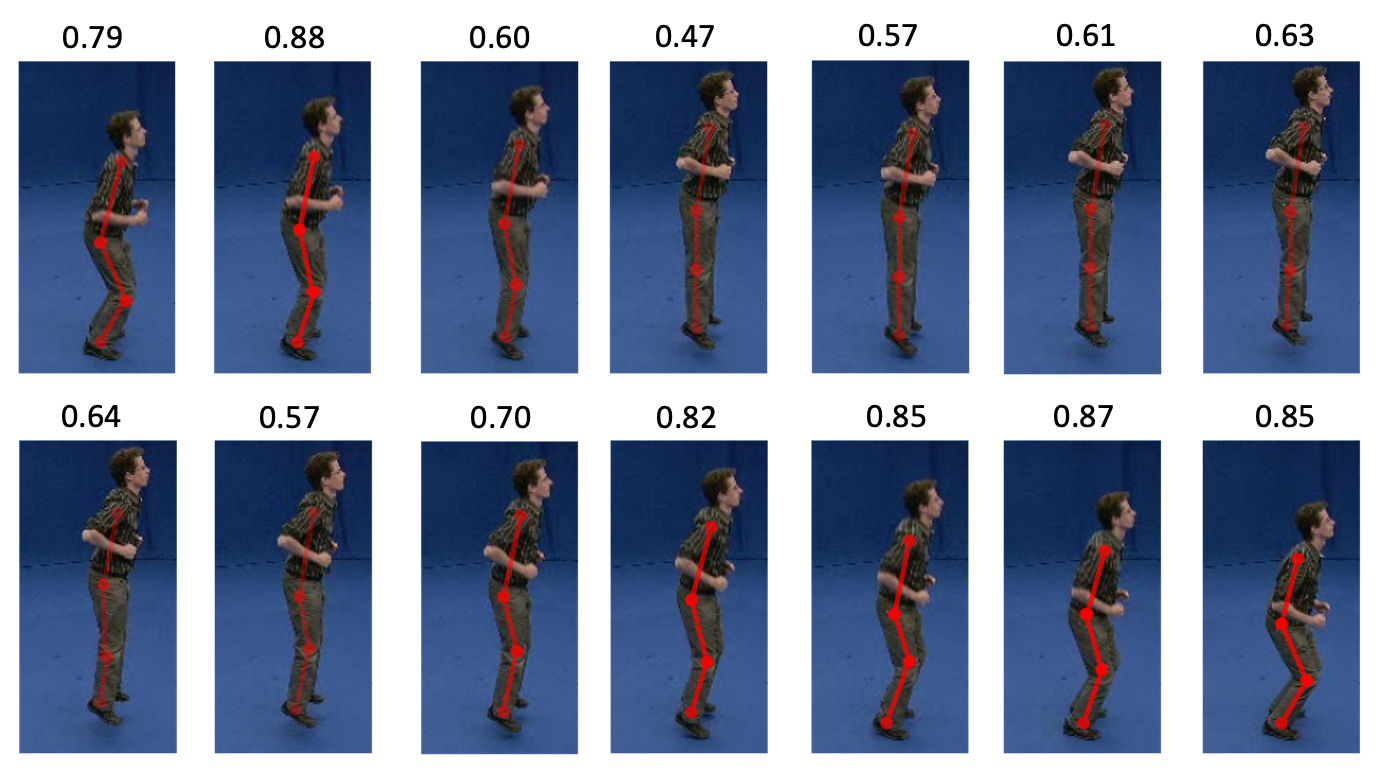}
    \caption{Self-attention of the [CLS] token, computed as the normalized sum of the last layer of the different heads. Scores give a direct insight into the frames exploited by AcT to produce the classification output. The example clearly shows how bending positions are more insightful for the network to predict the jumping-in-place action. In the image, the attention score defines the skeleton alpha channel.}
    \label{fig:attention_frames}
\end{figure}

\begin{figure}[ht]
    \centering
    \includegraphics[width=\columnwidth]{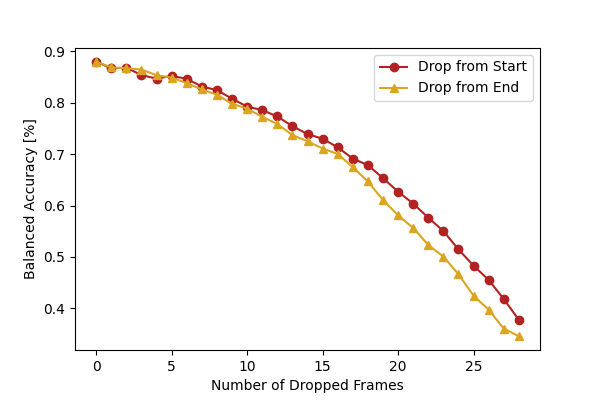}
    \caption{AcT-M balanced accuracy with an incremental reduction of temporal information. Due to the intrinsic nature of the network, it is possible to reduce the number of temporal steps without a retraining or any kind of explicit adaptation.}
    \label{fig:drop_frames}
\end{figure}

Instead, in Fig.\ref{fig:attention_frames}, the last-layer self-attention scores of the [CLS] token are shown together with the RGB and skeleton representations of the scene. The scores are computed as the normalized sum of the three attention heads and give a direct insight into the frames exploited by AcT to produce the classification output. It can be seen that bent poses are much more informative for the model to predict the jumping-in-place action.

Moreover, we analyze the behavior of the network under a progressive reduction of temporal information. That can be easily done without retraining due to the intrinsic nature of AcT. In Fig.\ref{fig:drop_frames}, we present how the test-set balanced accuracy is affected by frame dropping. The two curves show a reduction starting from the beginning and the end of the temporal sequence, respectively. It is interesting to notice how the performance of AcT degrades with an almost linear trend. That highlights the robustness of the proposed methodology and demonstrates the possibility of adapting the model to applications with different temporal constraints.

\begin{figure}[ht]
\centering
    \includegraphics[scale=0.65]{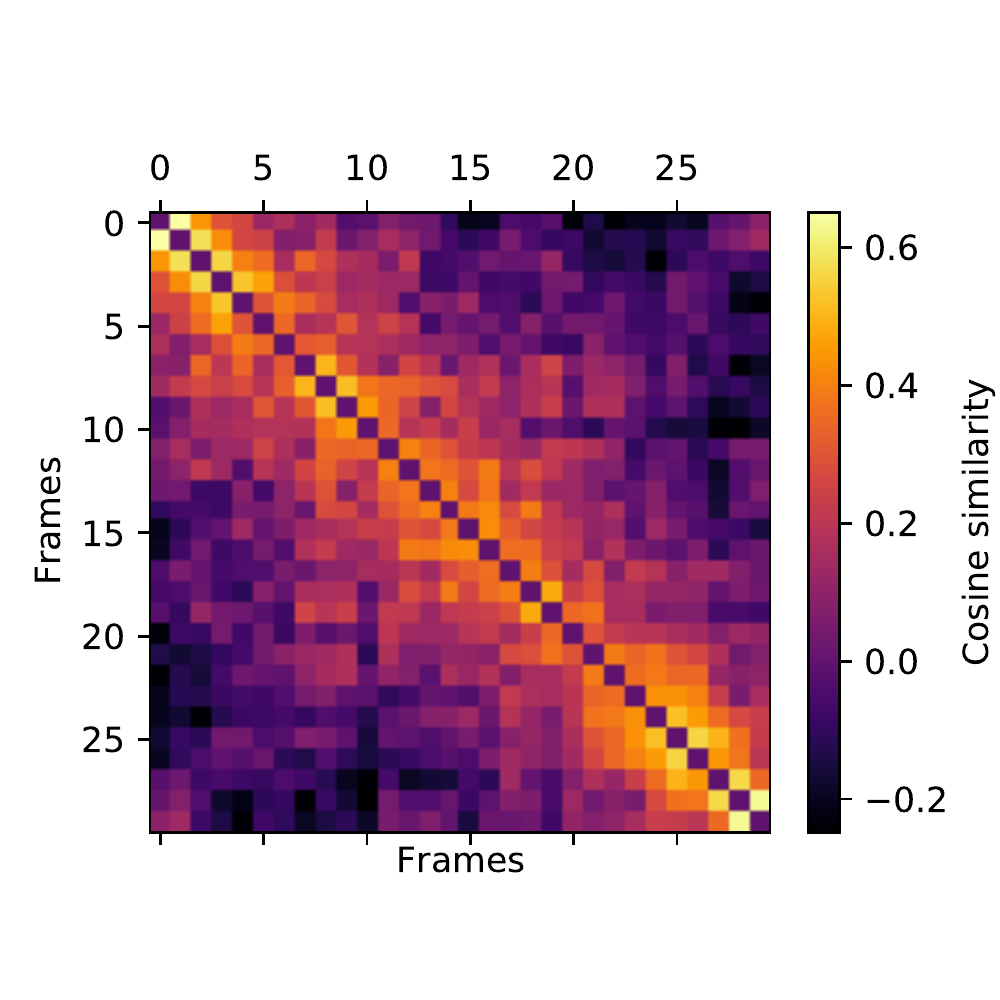}
    \vspace{-20pt}
    \caption{Cosine similarities of the learned $T$ position embeddings of AcT-M model.}
    \label{fig:cosine}
\end{figure}

Finally, we also study the positional embeddings of the AcT-M model by analyzing their cosine similarity, as shown in Fig.\ref{fig:cosine}. Very nearby position embeddings demonstrate a high level of similarity, and distant ones are orthogonal or in the opposite direction. This pattern is constant for all $T$ frames of the sequence, highlighting how actions are not particularly localized and that relative positions are essential for all the frames.

\subsection{Latency Measurements} % simo 
We test the performance of all the considered models for real-time applications. To do so, we use the TFLite Benchmark\footnote{https://www.tensorflow.org/lite/performance/measurement} tool, which allows running TensorFlow Lite models on different computing systems and collecting statistical results on latency and memory usage. In our case, two CPUs are employed to measure model speed both on a PC and a mobile phone: an Intel i7-9700K for the former and the ARM-based HiSilicon Kirin 970 for the latter. In both experiments, the benchmark executes 10 warm-up runs followed by 100 consecutive forward passes, using 8 threads.

\begin{figure}[ht]
    \centering
    \includegraphics[width=\columnwidth]{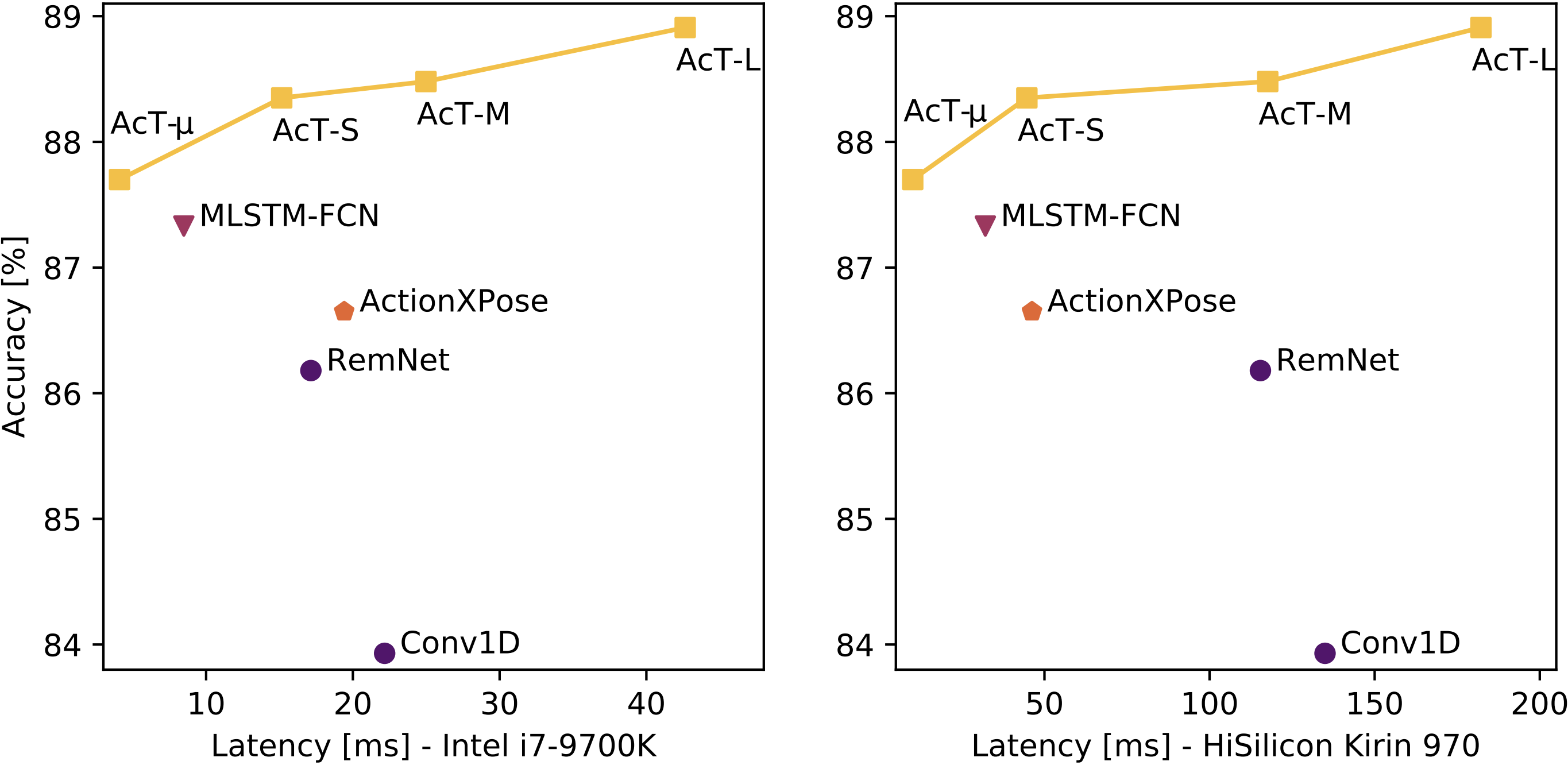}
    \caption{Study of the latency of different tested models on a high-performance Intel CPU and on a mobile phone equipped with an ARM-based CPU.}
    \label{fig:latency}
\end{figure}

The results of both tests are reported in Fig. \ref{fig:latency}, where only the MLP has been ignored because, despite being the fastest-running model, its accuracy results are much lower than its competitors. The graph shows the great computational efficiency of Transformer-based architectures, whereas convolutional and recurrent networks result in heavier CPU usage. Indeed, in the Intel i7 case, REMNet achieves almost the same speed as AcT-S, but its accuracy is 2\% lower. Moreover, AcT-$\mu$ is able to outperform REMNet, running at over four times its speed. MLSTM-FCN and ActionXPose, being smaller models, achieve lower latencies than the baselines: the former stays between AcT-$\mu$ and Act-S, while the latter performs similarly to AcT-S. Those results are remarkable but still outperformed by AcT-$\mu$ both on accuracy and speed.

The difference with the baselines is even more evident on the ARM-based chip, as convolutional architectures seem to perform poorly on this kind of hardware. Indeed, REMNet and Conv1D run as fast as AcT-M with significantly lower accuracies, and AcT-$\mu$ is ten times quicker. Nothing changes for what concerns MLSTM-FCN and ActionXPose, less accurate and three times slower than AcT-$\mu$.

%%%%%%%%%%%%%%%%%%%%%%%%%%%%%%%%%%%%%%%%%%%%%%%%%%%%%%%%%%%%%%%%%%%%%%%%%%%%%%%%
\section{CONCLUSION}
\label{conclusion}
In this paper, we explored the direct application of a purely Transformer-based network to human action recognition. We introduced the AcT network, which significantly outperforms commonly adopted models for HAR with a simple and fully self-attentional architecture. In order to limit computational and power requests, building on previous HAR and pose estimation research, the proposed methodology exploits 2D skeletal representations of short time sequences, providing an accurate and low latency solution for real-time applications. Moreover, we introduced MPOSE2021, a large-scale open-source dataset for short-time human action recognition, as an attempt to build a formal benchmark for future research on the topic. Extensive experimentation with our methodology clearly demonstrates the effectiveness of AcT and poses the basis for meaningful impact on many practical computer vision applications. In particular, the remarkable efficiency of AcT could be exploited for Edge AI, achieving good performance even on computationally limited devices. Future work may further investigate the AcT architecture with 3D skeletons and long sequence inputs. Moreover, relational information between points of the estimated pose are currently scarcely exploited by our methodology. So, future investigations may attempt to include skeleton graphs as prior knowledge in the positional embedding of the network without impacting latency.

%%%%%%%%%%%%%%%%%%%%%%%%%%%%%%%%%%%%%%%%%%%%%%%%%%%%%%%%%%%%%%%%%%%%%%%%%%%%%%%%

{\small
\bibliographystyle{ieee}
\bibliography{bibliography}
}
\end{document}